\documentclass[letterpaper]{article} 
\usepackage{aaai24}  
\usepackage{times}  
\usepackage{helvet}  
\usepackage{courier}  
\usepackage[hyphens]{url}  
\usepackage{graphicx} 
\usepackage{natbib}  
\usepackage{caption} 
\usepackage{algorithm}
\usepackage{algorithmic}
\usepackage{amsfonts}
\usepackage{amsmath}

\usepackage{newfloat}
\usepackage{listings}
\DeclareCaptionStyle{ruled}{labelfont=normalfont,labelsep=colon,strut=off} 
\floatstyle{ruled}
\newfloat{listing}{tb}{lst}{}
\floatname{listing}{Listing}
\newcommand{\system}{ViSTec}
\newcommand{\sharedthanks}{\thanks{Equal contribution; alphabetically ordered; each reserves the right to be listed first.}}
\newcommand{\correspondingauthor}{\thanks{Corresponding authors.}}
\title{ViSTec: Video Modeling for Sports Technique Recognition and Tactical Analysis}
\author {
    Yuchen He\sharedthanks,
    Zeqing Yuan\footnotemark[1],
    Yihong Wu,
    Liqi Cheng,
    Dazhen Deng\correspondingauthor,
    Yingcai Wu\footnotemark[2]
}
\affiliations {
    Zhejiang University\\
    \{heyuchen, leoyuan, wuyihong, lycheecheng, dengdazhen, ycwu\}@zju.edu.cn
}

\begin{document}

\maketitle

\begin{abstract}
The immense popularity of racket sports has fueled substantial demand in tactical analysis with broadcast videos. However, existing manual methods require laborious annotation, and recent attempts leveraging video perception models are limited to low-level annotations like ball trajectories, overlooking tactics that necessitate an understanding of stroke techniques. State-of-the-art action segmentation models also struggle with technique recognition due to frequent occlusions and motion-induced blurring in racket sports videos. 
To address these challenges, We propose \system{}, a \textbf{Vi}deo-based \textbf{S}ports \textbf{Tec}hnique recognition model inspired by human cognition that synergizes sparse visual data with rich contextual insights. Our approach integrates a graph to explicitly model strategic knowledge in stroke sequences and enhance technique recognition with contextual inductive bias. A two-stage action perception model is jointly trained to align with the contextual knowledge in the graph. 
Experiments demonstrate that our method outperforms existing models by a significant margin. Case studies with experts from the Chinese national table tennis team validate our model's capacity to automate analysis for technical actions and tactical strategies. More details are available at: https://ViSTec2024.github.io/.
\end{abstract}

\section{Introduction}
Racket sports, including tennis, badminton, and table tennis, are distinguished by their highly strategic nature, drawing millions of players and fans to explore in-depth tactical analysis. A game in racket sports consists of rallies, which are sequences of strokes executed by players alternately from both sides. Here, a \textbf{stroke} refers to the action of hitting the ball with a racket, and each stroke can employ a certain \textbf{technique}, such as ``topspin'' and ``push.'' A \textbf{tactic} is characterized by a series of consecutive stroke techniques.
In a match, it is of paramount importance to analyze the interplay of tactics and players' characteristics in technical actions. Both aspects necessitate an understanding of the techniques used in each stroke. 
Every year, numerous tournaments are held and broadcasted, generating a vast amount of video data. Therefore, modeling broadcast videos on the technique level is a promising direction for facilitating and democratizing racket sports analysis.

Previous methods for racket sports analysis~\cite{DBLP:journals/tvcg/WuLSJZWZ18, tennivis2014tom, tivee2022chu} heavily depend on fine-grained data and suffer from low scalability due to the demand of labor-intensive annotation from domain experts. 
On the other hand, recent attempts utilize video perception models for automatic annotation, yet are hindered by the sparsity of visual information in broadcast videos, particularly challenges such as motion-induced blurring, subtle movement amplitude, and frequent occlusions of wrists and rackets. Moreover, they concentrate only on low-level objects and coarse-grained events, such as ball trajectories~\cite{DBLP:conf/avss/HuangLCIP19} and stroke timestamps~\cite{DBLP:conf/cvpr/VoeikovFB20}. 
As for techniques, which involves high-level semantics and contextual knowledge, it cannot be taken as mere actions and state-of-the-art action segmentation models~\cite{DBLP:journals/corr/abs-2207-12730} fall short in recognizing racket sports techniques.

In this paper, we aim to recognize and analyze fine-grained stroke techniques from low-quality broadcast videos, bridging the gap between professional expertise and automated analysis.
We propose \system{}, which incorporates domain knowledge as inductive prior.
We select table tennis as a representative racket sport for our study, considering it the most challenging for stroke recognition and well-known for being highly strategic. We collaborated closely with two senior data analysts from the Chinese national table tennis team when developing and evaluating our methods.

\system{} is composed of an action perception module and a domain knowledge module.
The action perception module operates in a two-stage manner, leveraging visual information. It first segments each stroke clip from raw video, and then classifies the specific stroke techniques, working collaboratively with the domain knowledge module to enhance accuracy.
The domain knowledge module models contextual knowledge, with a focus on technique sequence dependencies. It adopts the form of a graph to explicitly represent the transition relations between techniques, thus integrating this relational understanding as prior knowledge.
The two modules are thoughtfully aggregated and jointly trained to achieve better synergy.

To demonstrate the effectiveness of our framework, we perform comparative experiments with state-of-the-art action segmentation models and conduct an ablation study to examine individual components.
Furthermore, we conduct case studies on the 2022 Table Tennis World Cup to analyze players' playing styles and optimal strategies under different circumstances.
The results demonstrate that our model exhibits a proficient understanding of stroke techniques, enabling automatic tactical analysis.

The contributions of this paper are as follows:
\begin{itemize}
    \item We address the problem of video-based technique recognition in racket sports, facilitating automatic tactical analysis.
    \item We propose a novel framework that leverages both sparse visual information and contextual domain knowledge for video understanding, achieving state-of-the-art performance in sports technique recognition. 
    \item We conduct experiments and case studies to demonstrate the usefulness of our model and obtain valuable insights validated by professional analysts.
\end{itemize}

\section{Related Work}

\subsection{Sports Action Recognition and Segmentation}
Action recognition aims to identify the categories of human actions in videos.
Existing studies~\cite{DBLP:conf/cvpr/KarpathyTSLSF14, 
 DBLP:conf/iccv/TranBFTP15, DBLP:conf/cvpr/CarreiraZ17, DBLP:journals/pami/0002X00LTG19} propose a series of neural network architectures to learn action representations from raw video or optical flow.

For example, Carreira and Zisserman~\cite{DBLP:conf/cvpr/CarreiraZ17} proposed a two-stream inflated 3D convolutional network architecture as a backbone video model.
However, action recognition focuses on video-level classification, thereby failing to analyze long videos containing multiple actions.

Researchers further delve into action segmentation, a task that involves not only recognizing actions but also localizing the time intervals they occur.
There are several methods to obtain time segments, such as sliding windows~\cite{DBLP:conf/eccv/KimKK22}, proposal generation~\cite{DBLP:conf/eccv/LinZSWY18, DBLP:conf/iccv/LinLLDW19}, and per-frame labeling~\cite{DBLP:conf/cvpr/ShouCZMC17}.
For sports analysis, datasets are the key. Liu et al.~\cite{fineaction2022liu} proposed a dataset that covers a series of actions across different sport types, and evaluated BMN~\cite{DBLP:conf/iccv/LinLLDW19}, DBG~\cite{lin2020fast}, G-TAD~\cite{xu2020g}.
Finegym~\cite{DBLP:conf/cvpr/ShaoZDL20a} is a comprehensive gymnastic dataset with 530 action types.
$P^2A$~\cite{DBLP:journals/corr/abs-2207-12730} proposes a fine-grained table tennis dataset and assesses a series of localization and recognition models.
These datasets open up a research direction of recognizing complex and dynamic sports actions, but state-of-the-art methods fail on racket sports scenarios with challenges such as frequent occlusions and subtle movements.

\subsection{Racket Sports Data Mining and Analysis}
The popularity of racket sports has garnered the interest of data mining.
For tennis analysis, a series of studies analyze and visualize the scoring outcome~\cite{tennivis2014tom} and ball trajectories~\cite{courttime2020tom}.
Some research attempts have been devoted to synthesizing or reconstructing player actions~\cite{zhang2021vid2player, tennis2023zhang} from broadcast videos.
For badminton, researchers have employed AR/MR technologies to visualize and analyze 3D shuttle trajectories~\cite{shuttle2021ye, tivee2022chu, lin2023vird}.
For table tennis, a series of visual analytics systems, such as iTTVis~\cite{DBLP:journals/tvcg/WuLSJZWZ18} and Tac-Simur~\cite{DBLP:journals/tvcg/WangZDCXZZW20}, are developed to analyze the attributes between consecutive strokes.
Tac-Valuer~\cite{DBLP:conf/kdd/WangDXSHCZZZW21} combines deep learning and abductive learning~\cite{abd2019dai} to incorporate sequence dependency into the stroke classification.
However, these methods rely on fine-grained attributes that are labeled by the experts manually.
To improve data accessibility, EventAnchor~\cite{DBLP:conf/chi/DengWWWXZZZW21} combines computer vision models and human-computer interaction techniques to improve data annotation efficiency.

To further enable tactical analysis, we develop a method to recognize high-level stroke techniques from broadcast videos.
The results can be directly used to analyze the player tactics without additional data annotation, democratizing large-scale analysis on tactics.

\begin{figure*}[!htb]
    \centerline{\includegraphics[width=\linewidth]{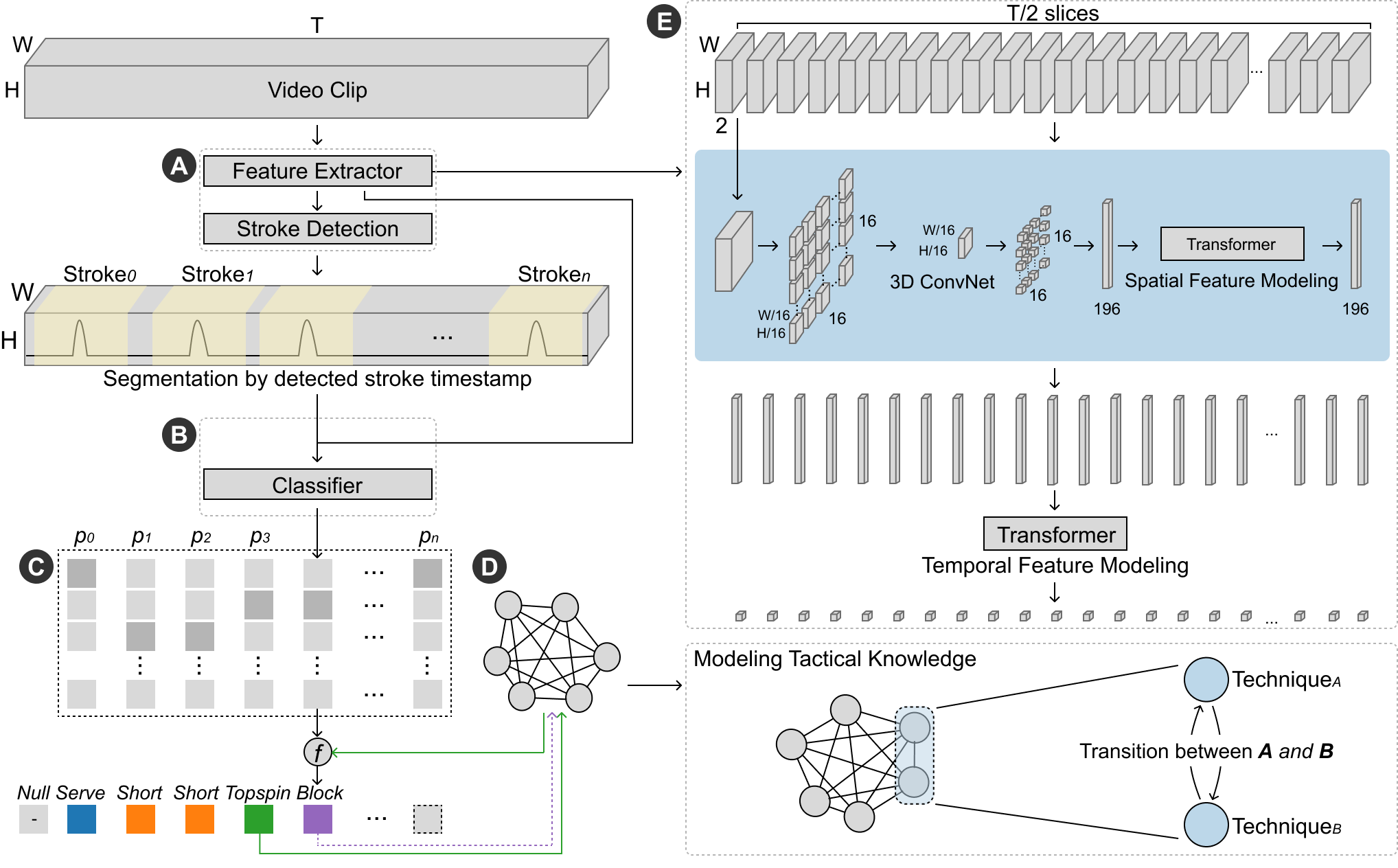}}
    \caption{The framework of \system{}. (A) is the stroke segmentation module. (B) is the $cls$ module, with segmented stroke features as input and probability distributions for each segment as output, as shown in (C). (D) is the $grh$ module for domain knowledge modeling. (E) shows the detail of video feature extractor.}
    \label{fig:framework}
\end{figure*}

\section{Data Descriptions and Notations}

\subsection{Racket Sports Data}
In racket sports, tactical analysts usually focus on fine-grained stroke techniques.
The stroke techniques refer to specific types of action to wave the racket and hit the ball. 
Different stroke techniques result in various ball spins and moving directions, which consequently affect the ball trajectory to a great extent. 
Taking table tennis as an example, strokes can be categorized to eight techniques, such as serve, topspin, short, and block. Additional attributes can be incorporated to provide a more fine-grained classification, such as forehand and backhand.
In tennis, stroke techniques can also be categorized into eight types~\cite{tennis2023zhang}.
In addition, in racket sports analysis, consecutive three strokes are usually considered to be a minimal unit for tactics~\cite{tacminer21wang}.
Professional analysts normally conduct analysis at tactic or stroke level~\cite{DBLP:journals/tvcg/WangZDCXZZW20}.

\subsection{Notations}
We first introduce the problem and notations used in this study.
In our scenario, the input is a video $V=\{v_1, v_2,...,v_T\}$ with $T$ frames and the output is the sequence $S=\{(s_1,t_1), (s_2, t_2), ..., (s_N,t_N)\}$ with $N$ strokes.
The value $s_i$ represents the stroke technique and $t_i$ is the timestamp of the stroke.
In table tennis, for example, the $s_i$ can be techniques such as topspin, serve, short, block, push, flick or smash.
It is noted that compared to ordinary action segmentation tasks where $t_i$ is an interval lasting for frames or seconds, in racket sports the interval of the action is usually ambiguous.
Therefore, researchers represent the stroke action as an immediate event and only record the moment when the ball hits the racket as the time of stroke event~\cite{DBLP:conf/cvpr/VoeikovFB20}.

\section{\system{}}

In this section, we introduce our two-stage framework for stroke recognition. The framework is demonstrated in Fig.~\ref{fig:framework}.

\subsection{Video Feature Modeling}
We first model the spatial and temporal features of videos.
Given that stroke is fast-paced and lasts for only several frames, it is required to extract frame-wise features~\cite{DBLP:conf/cvpr/0001WL022}.
We employ a transformer-based structure for frame-wise feature extraction.

Because of the success of vision transformers in the tasks of video modeling, we select VideoMAE~\cite{DBLP:journals/corr/abs-2203-12602} as our backbone.
VideoMAE first divides a long video clip into slices and extracts features slice by slice with a slice length of 16 frames.
However, in our scenario, every frame counts because of the short durations of table tennis strokes.
Therefore, we use a slice length of 2 frames.
A slice is in the shape of $H\times W\times 2$ ($H=224, W=224$), and each slice is forwarded into a 3-dimensional convolutional layer with the filter shape of $H/16\times W/16\times 2$ and stride shape of $H/16\times W/16\times 2$.
As a result, a slice is converted into $16\times 16$ patches of features.
Each patch represents the spatial feature of a region in the original slice.
All patches of features are then concatenated together and forwarded into a transformer-based network.
The network can model the relations between feature patches and construct a spatial feature of the video slice.

We further model the temporal feature of the whole video based on the spatial feature of each video slice.
After the spatial modeling, a feature of $T/2\times 196\times 1280$ is obtained.
The feature is average pooled over the patch dimension and transformed into the shape of $T/2\times 1280$.
The feature is then forwarded into a multi-layer transformer encoder to model the temporal features.
Then the temporal features are forwarded into a fully-connected layer to predict frame-wise stroke attributes, which are represented as multi-class distributions of size $T/2\times C$.

\subsection{Stroke Segmentation}\label{sec:stroke_segmentation}

With the feature extracted from the backbone model, we perform stroke segmentation $seg$ with a fully-connected network, predicting the probability of each timestamp.
\begin{equation}
    seg(V)=\hat{\mathbb{P}}(t_1, t_2, ..., t_T|V)
\end{equation}
, where $\hat{\mathbb{P}}$ is the predicted probability of a stroke given the video $V$. 
Compared to other segmentation models~\cite{DBLP:conf/iccv/LinLLDW19, xu2020g} that focus on actions with relatively longer durations (lasting for seconds or minutes), we model strokes to be instant events.
Therefore, the $seg$ predicts a series of signals of the stroke probability at each timestamp.
During training, we convert stroke events into a series of cosine signals, where the peak is the moment of ball hitting~\cite{DBLP:conf/cvpr/VoeikovFB20}.
\begin{equation}
  \mathbb{P}(t_i) =
    \begin{cases}
      cos\frac{(t_i-t_s)\pi}{\sigma} & \text{if $|t_i-t_s|\leq \frac{\sigma}{2}$}\\
      0 & \text{otherwise}
    \end{cases}    
\end{equation}
where $t_s$ is the stroke timestamp which is the closest to the target timestamp $t_i$. In this study, we use a $\sigma=8$ because the strokes have an average length of about eight frames.

During training, we evaluate the difference between the predicted probability $\hat{\mathbb{P}}$ and target probability $\mathbb{P}$ using binary cross entropy $l_{BCE}$.
\begin{equation}
    l_{BCE}(\hat{\mathbb{P}}, \mathbb{P})=\frac{1}{T}\sum_{i=1}^{T}\hat{p_i}\cdot p_i+(1-\hat{p_i})\cdot (1-p_i),
\end{equation}
where $\hat{p_i}=\hat{\mathbb{P}}(t_i|V)$ and $p_i=\mathbb{P}(t_i)$.
For a video clip, the differences of all timestamps are computed and then reduced with the mean value.

With the results from the stroke segmentation $seg$ module, we further filter the timestamps with high probabilities as the stroke segments.
To avoid ragged segments, we merge the strokes between which the temporal distance is smaller than $\frac{\sigma}{2}$ frames.
As a result, the predicted stroke timestamps $\hat{t}_i(1\leq i\leq N)$ are intervals.

\subsection{Stroke Classification with Contextual Knowledge}
After segmentation, we further classify the strokes into fine-grained techniques using a classification module $cls$ and a graph module $grh$.
The $cls$ is a fully-connected network that takes the segmented stroke features as input and predicts the technique types.
\begin{equation}
    \hat{s}_i=cls(f_i)=\mathbb{P}(tec|f_i),
    \label{equation:cls}
\end{equation}
where $f_i$ is the aggregated features of the stroke $\hat{s}_i$ by the time interval $\hat{t}_i$.
The result $\mathbb{P}(tec|f_i)$ is the distribution of different techniques.

Predicting stroke techniques solely from visual features might fail in distinguishing strokes that are visually similar to each other.
Naively selecting the best-predicted technique for each stroke might result in invalid stroke sequences.

\textbf{Contextual Knowledge Learning.} Considering the intricacies of table tennis stroke techniques, which require contextual inference for proper interpretation, we introduce the $grh$ module, a graph-based data structure to model the contextual information of table tennis game videos.

As illustrated in Figure \ref{fig:framework}, $grh$ represents a directed graph comprised of several nodes and edges, denoted as
\begin{equation}
    G_{grh} = (V_T, E),
\end{equation}
where the node set is defined as $V_T = \{v_{tec_0}, \dots, v_{tec_{m-1}}\}$, with each node $v_{tec_k}$ symbolizing the classification label $tec_k$ of a distinct stroke technique.
Notably, a designated ``null'' node is utilized to represent an empty label, which serves its purpose when generating sequences, as the initial label in a sequence.
The edge set $E = \{e_0, ..., e_{m\times(m-1)-1}\}$ encompasses all possible directed edges connecting pairs of nodes except edges from all nodes to ``null'' node, and for each edge from $v_{tec_A}$ to $v_{tec_B}$, there exists a weight representing the transition from stroke technique $tec_A$ to $tec_B$.

\textbf{Joint Training of $cls$ and $grh$.} To incorporate contextual information, our model departs from the approach in Eq.~\ref{equation:cls}, where aggregated features $f_i$ were the sole input. Instead, we introduce the preceding stroke's label along with the aggregated features of the current stroke as input. When the preceding stroke is absent, as in the case of the first stroke, the label ``null'' is employed to denote the preceding stroke.

Given the knowledge of the previous stroke's label, denoted as $tec_p$, we leverage this label to locate the corresponding node within $grh$. Subsequently, by querying the directed edges emanating from this node, we derive a weight vector representing transitions from label $tec_p$ to next possible labels. The length of this vector equals the number of distinct technique labels, with its shape aligning with the output of the classification model. In the subsequent discourse, we shall refer to such weight vectors as $W_{tec_p}$.

We update the parameters of the classification model as follows. Initially, for the model's output that hasn't undergone normalization, we conduct min-max normalization to ensure all values within the vector are confined to the $[0,1]$ range. Subsequently, we obtain the transition weight vector from $grh$, similarly ensuring that all values within the vector fall within the $[0,1]$ range. We then combine the two vectors with a specific proportion. We compute the cross-entropy loss and execute backpropagation for parameter updates.
\begin{equation}
    l_{CE}(\mathbb{P}_c, \hat{\mathbb{P}}_c) = -\sum_{i} p_{c_i} \cdot \hat{p}_{c_i}
\end{equation}

\begin{equation}
    \hat{\mathbb{P}}_c = Softmax(MinMax(cls(f_i))+\alpha W_{tec_p}), \label{equation:final_probability}
\end{equation}
where $\hat{\mathbb{P}}_c$ and $\mathbb{P}_c$ are the predicted and target classification probability respectively, $\hat{p}_{c_i} = \hat{\mathbb{P}}_c(tec_i|f)$, $p_{c_i} = {\mathbb{P}}_c(tec_i)$, $\alpha$ is a hyperparameter for the combination.

When updating the weights of the $cls$ module, it is equally essential to update the weights within the graph ($grh$). In \system{}, these weights are updated with an adaptive stride. It is noteworthy that at the commencement of training, the graph is initialized using all known technique sequences from the training set. This initialization enables the weights to roughly reflect the transition probabilities between various pairs of techniques observed in the training data.

Following a single forward pass through the model, we obtain $cls(f_i)$, allowing us to compute $\hat{\mathbb P}_c$ as indicated in Eq.~\ref{equation:final_probability}. This estimation is then used to update the edge weights $W_{tec_p}$ originating from the node corresponding to the previous stroke label $tec_p$ within $grh$.

Should $\hat{\mathbb P}_c$ deduce the correct label, i.e., if the label with the highest confidence in $\hat{\mathbb P}_c$ aligns with the ground truth label, there is no need to update the edge weights of $grh$. However, if discrepancies arise, we adopt the following strategy for updates. Let the predicted label inferred from $\hat{\mathbb P}_c$ be denoted as $tec_{pred}$, and the ground-truth label as $tec_{gt}$. Our objective is to diminish the transition $tec_p \rightarrow tec_{pred}$ within $W_{tec_p}$ while reinforcing the transition $tec_p \rightarrow tec_{gt}$.
\begin{algorithm}[!htb]
	\renewcommand{\algorithmicrequire}{\textbf{Input:}}
	\renewcommand{\algorithmicensure}{\textbf{Output:}}
	\caption{Updating $W_{tec_p}$}
	\label{algorithm: updating}
	\begin{algorithmic}[1]
		\REQUIRE
			Weight vector $W_{tec_p}$,
			predicted label of current segment $tec_{pred}$,
               and ground-truth label $tec_{gt}$.
		\ENSURE
			Updated weight vector $W^\prime_{tec_p}$.
        \STATE Initialize: $W^\prime_{tec_p} \leftarrow W_{tec_p}$\\
        \STATE $W^\prime_{tec_p}[tec_{pred}] \leftarrow (1-\beta U(cls(f_i)))W^\prime_{tec_p}[tec_{pred}]$\\
        \STATE $W^\prime_{tec_p}[tec_{gt}] \leftarrow (1+\beta U(cls(f_i)))W^\prime_{tec_p}[tec_{gt}]$\\
        \STATE Normalization: $W^\prime_{tec_p} \leftarrow W^\prime_{tec_p}/\max(W^\prime_{tec_p})$
	\end{algorithmic}
\end{algorithm}

The hyperparameter $\beta$ controls the stride of the update. $U(cls(f_i))$ constitutes a crucial element in the adaptive update stride and effectively incorporates the uncertainty of the classification confidence of $cls$ concerning input features. Here, we employ entropy to quantify the uncertainty in the classification confidence provided by $cls$. The computation of $U(cls(f_i))$ is as follows:
\begin{equation}
    U(cls(f_i)) = (1 - \frac{Entropy(Softmax(cls(f_i)))}{\mu}),
    \label{equ:uncertainty}
\end{equation}
where $\mu$ represents the maximum entropy of the probability distribution corresponding to the number of classes in the classification. It can be precomputed and regarded as a constant in training.

The significance of determining the update stride in this manner lies in the fact that when $cls$ assigns an incorrect label with lower uncertainty, it indicates a more serious model error at that moment. Consequently, there is a greater need for rectifying the edge weights of $grh$ to assist the overall model in making accurate classification. On the other hand, during the early stages of training, the model's classification uncertainty is relatively high. Consequently, the update stride is smaller, which serves to prevent rapid disruption of the initial graph weights. This approach enhances the model's robustness during the initial training phase.

\subsection{Inference of \system{}}

As depicted in Figure \ref{fig:framework}, \system{}'s inference process contains two stages. First, using the $seg$ module, we acquire several time intervals $\hat{t}_i (1\leq i\leq N)$ representing different strokes from the input match video. Centered around each of these intervals, we extend the respective time spans to form segments, each spanning no more than 40 frames. When adjacent extended intervals overlap, we designate the midpoint between them as the separator for creating distinct segments. Through this segmentation, we ensure that every segment encompasses only a single stroke, each frame belongs to a solitary segment excluding redundant frames, such as those capturing player preparation before a serve.

Second, we sequentially iterate each segment and employ the features of the current segment along with the predicted label of the previous segment (or a "null" label if there is no previous segment) as inputs. By utilizing Eq.~\ref{equation:final_probability}, we compute the predictive probability distribution and choose the label with the highest confidence as the prediction. Upon completing this inference iteration, we obtain the sequence of stroke techniques for the given match video.

\section{Experiments}
\subsection{Experiment Dataset}
All experiments are performed on a dataset constructed from broadcast videos of World Table Tennis (WTT) games.
We use table tennis as an experimental scenario because it is the most challenging racket sport for video analysis considering the frequent occlusion, minimal movement amplitude, and blurring caused by the rapid pace.
We collected 4000 rally clips segmented from 18 games by recognizing scoreboard changes~\cite{DBLP:conf/chi/DengWWWXZZZW21}.
Each clip includes a series of strokes.
We labeled the timestamps of each stroke to train the stroke segmentation module.
However, the labeling of the stroke techniques requires professional table tennis knowledge and experience.
Therefore, we consulted with professional athletes who have been members of provincial teams or national reserve teams.

\begin{table}[!htb]
\centering
\begin{tabular}{l|cccccc}
    \hline
    Models & F1@\{10,25,50\} & Acc. & Edit\\\hline
    C2F-TCN & 50.8, 45.3, 32.0 & 61.1 & 45.0\\
    ASFormer & 75.2, 73.3, 69.7 & 77.5 & 73.7\\
    UVAST & 75.2, 74.3, 71.3 & 76.1 & 74.1\\
    SSTDA & 76.0, 73.2, 67.0 & 76.5 & 72.5\\
    MS-TCN & 76.8, 74.8, 71.1 & 78.2 & 73.9\\\hline
    \system{} w/o $grh$ & 76.3, 76.2, 75.3 & 82.0 & 74.3\\
    \system{} w/o $U$ & 77.9, 77.7, 77.0 & 82.2 & 74.8\\\hline
    \system{} & \textbf{79.3}, \textbf{79.2}, \textbf{78.5} & \textbf{83.5 }& \textbf{76.3}\\\hline
\end{tabular}
\caption{Experiment results including ablation studies of the proposed method and baselines.}
\label{table2}
\end{table}

\subsection{Comparative Study}
We compare \system{} with state-of-the-art action segmentation models.
Specifically, we train C2F-TCN~\cite{singhania2021coarse}, ASFormer~\cite{DBLP:conf/bmvc/YiWJ21}, UVAST~\cite{DBLP:conf/eccv/BehrmannGKGN22}, SSTDA~\cite{DBLP:conf/cvpr/ChenLBAK20} and MS-TCN~\cite{DBLP:conf/cvpr/FarhaG19} on the table tennis dataset.
Noted that these models rely on visual features that are extracted with backbone models, such as I3D~\cite{DBLP:conf/cvpr/CarreiraZ17}.
However, I3D performs pooling along the temporal dimension, which makes it inappropriate to process the table tennis dataset, where a stroke action only lasts for several frames.
Therefore, to ensure a fair comparison, we extract frame-wise features using the same backbone as \system{} for the baseline models. Moreover, we generate frame-wise labels from ground-truth annotations, which consist of sequences of timestamp-technique pairs. Centered around each timestamp, we assign corresponding technique labels to frames within a range of no more than 40 frames. In cases where adjacent intervals overlap, we separate them by the midpoint between them. Frames that remain unassigned labels are considered background frames.

We adopted evaluation metrics commonly employed in the field of action segmentation~\cite{DBLP:conf/cvpr/LeaFVRH17}. As illustrated in Table \ref{table2}, ``Acc.'' corresponds to the frame-wise accuracy, ``Edit'' denotes the segmental edit score, and ``F1@$\{10,25,50\}$'' signifies the segmental F1 score at overlapping thresholds of $10\%$, $25\%$, and $50\%$, respectively. Benefiting from our two-stage design, we attain notable segmentation results in the first stage, while leveraging domain knowledge to enhance classification accuracy in the second stage. Our proposed approach excels across these evaluation metrics, achieving state-of-the-art results.

\subsection{Ablation Study}
We evaluate the effectiveness of different modules by removing them. Initially, we eliminated the $grh$ module to investigate its impact on the model's performance. Specifically, in the second stage, we solely employed the $\mathbb{P}(tec|f_i)$ calculated from the output $cls(f_i)$ in Eq.~\ref{equation:cls} to train the classification model by cross-entropy loss. As indicated in Table \ref{table2}, the performance metrics of \system{} without the $grh$ module were consistently inferior to those of the complete \system{}, indicating that the $grh$ module contributes positively to the model's performance.

In another experiment, we excluded the uncertainty term from the $grh$ update stride, which corresponds to the $U(cls(f_i))$ term defined in Eq.~\ref{equ:uncertainty}. This means the updating stride of the $grh$ is fixed. As shown in Table~\ref{table2}, the performance metrics of \system{} without the $U$ term were lower than those of the full \system{}, yet higher than those of \system{} without $grh$. This observation underlines that introducing the uncertainty term for dynamically updating the graph's weights enhances the model's performance.

\subsection{Qualitative Evaluation}
\begin{figure}[!htb]
\centerline{\includegraphics[width=\linewidth]{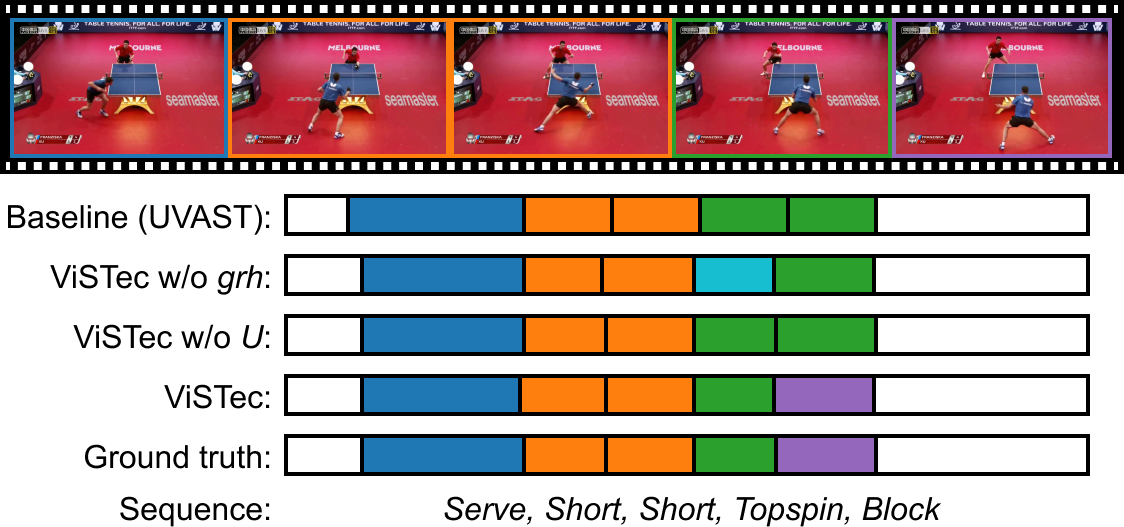}}
\caption{Illustration of segmentation result for a sample from our dataset with the ground truth sequence ``Serve, Short, Short, Topspin, Block''.}
\label{fig:segment}
\end{figure}

Figure~\ref{fig:segment} presents the segmentation results of a sample, comparing baseline (UVAST), \system{} without $grh$, \system{} without $U$, \system{}, and the ground truth. Notably, our method yields superior results in terms of segmentation and classification.
First, the start and end points predicted by the baseline model can not align well with the ground truth.
The baseline model tends to segment labels into equal-length segments except ``Serve''.
Differently, our proposed method detects the stroke event first and then performs the segmentation, thus allowing detecting actions with varying duration.
Second, our method demonstrates commendable classification performance as well.
The use of uncertainty and the graph module can effectively introduce domain knowledge learned from historic data to fix incorrect predictions.

Furthermore, offline tests on a single A100 GPU show \system{} achieving an inference speed of 39.3 frames per second, which exceeds the typical frame rate of broadcast match videos, enabling real-time processing.

\section{Evaluation with Case Studies}
To validate the effectiveness of \system{}, we present two case studies conducted with senior analysts from the Chinese table tennis team. The two cases address the most critical and complex facets of sports analysis that require domain knowledge: technique analysis and tactical analysis. Technique analysis demands meticulous observation of each technical action of different players, while tactical analysis necessitates careful attention to temporal correlations across various scales. In the first case, we analyze the player's technical actions based on visual features extracted from the video, uncovering the personalized characteristics and correlations. Second, we perform analysis based on sequences of stroke techniques obtained by \system{} from video, identifying tactics with a high scoring rate.

\textbf{Case 1: Analyzing Personalized Characteristics of Technical Actions}

Players have unique features in their technical actions and understanding the relation among technical actions is key to comprehending a player's characteristics. The conventional approach necessitates domain experts to review long videos and summarize various techniques manually. However, with the visual features extracted by our model, this process can now be accomplished automatically, streamlining the analysis and reducing reliance on manual expertise. 

\begin{figure}[!htb]
\centerline{\includegraphics[width=\linewidth]{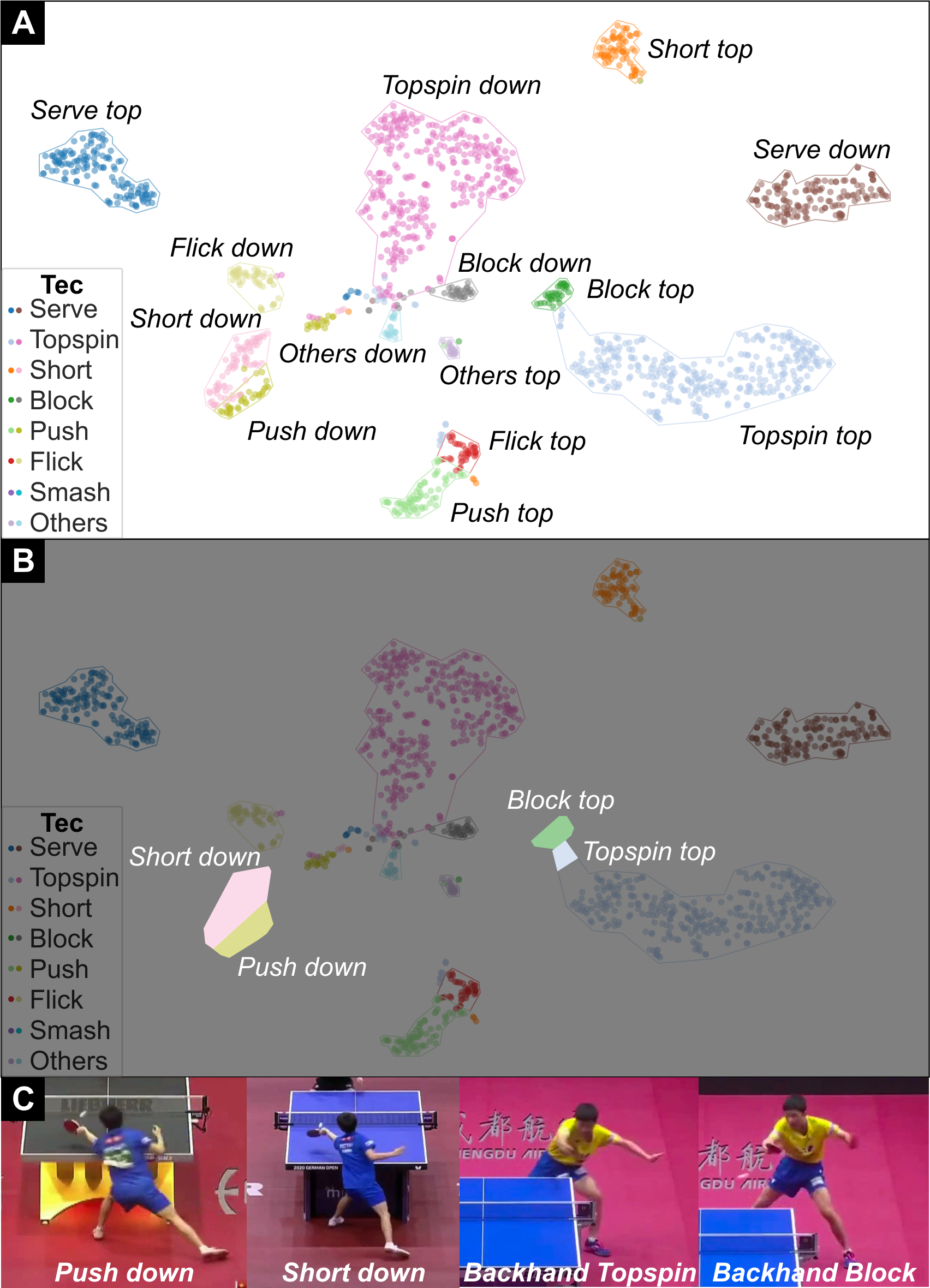}}
\caption{Case 1: (A) displays visual features of the strokes from two Japanese players with t-SNE. (B) highlights techniques that share noticeable similarities. (C) shows the technique actions highlighted in (B).}
\label{fig:case_1}
\end{figure}

After extracting features for each stroke from broadcast video with \system{}, we employ t-SNE for dimensionality reduction, projecting them onto a two-dimensional plane shown in a scatter plot in Figure~\ref{fig:case_1} (A). Notably, the stroke features form clusters based on technique categories, implying that \system{} has good observation on technical detail and context. This is especially impressive considering mere actions are often not informative enough to classify owing to occlusion and limited movement amplitude, making contextual details such as ball trajectory necessary.

As shown in Figure~\ref{fig:case_1} (B), for Japanese players, technique ``Block'' and ``Topspin'' exhibits striking analogy, as do ``Push'' and ``Short''. These similarities within deep visual features \textit{``reveal the high consistency and deceptive nature of their certain techniques,''} noted by the experts. This observation furnishes valuable insights, allowing opponents to enhance their preparation and anticipation of the players' moves in specific techniques, a critical factor in the fast-paced world of racket sports.
Similar analysis can be transferred to other players in real time, enabling the understanding of the unique characteristics of their opponent's actions.

\textbf{Case 2: Discovering Optimal Tactical Choices}

In racket sports, tactics hold paramount importance and are often meticulously selected, taking into account the opponents' characteristics, the current status of the game, and the individual's strengths. A tactic in table tennis refers to the techniques employed in consecutive strokes. Analyzing these tactics presents a complex challenge, given the degree of freedom in the temporal dimension and the multitude of possibilities. Traditionally, this analysis has required domain experts to identify and analyze the techniques used in each round, a task that can be demanding and time-consuming. Our proposed model, however, represents a significant advancement in this domain, capable of accurately recognizing sequences of techniques directly from raw video data, thereby unlocking new potentials in tactical analysis. We take 18 match videos from WTT to analyze tactical patterns of high scoring rate.

\begin{figure}[!htb]
\centerline{\includegraphics[width=\linewidth]{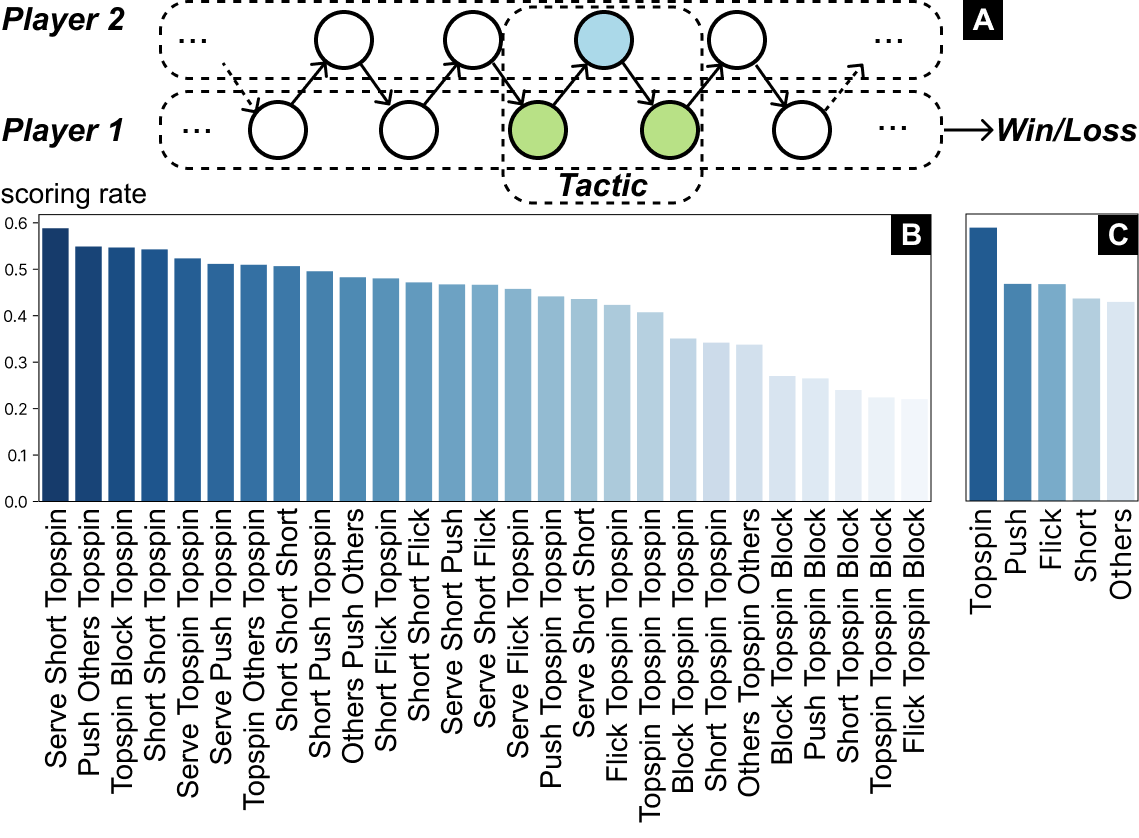}}
\caption{Case 2: (A) shows the structure of a table tennis tactic, consisting of consecutive three strokes. (B) illustrates the scoring rate of consecutive three strokes. (C) illustrates the scoring rate using different techniques after ``Serve, Short''.}
\label{fig:case_2}
\end{figure}

We begin by extracting sequences of stroke techniques from match videos using our model, subsequently conducting an analysis to discern the correlation between tactics and scoring rates in sets of three consecutive strokes. As demonstrated in Figure~\ref{fig:case_2}(B), the sequence ``Serve Short Topspin'' exhibits the highest scoring rate. This suggests that when serving on our side and the subsequent opponent's stroke involves a ``Short'' technique, the optimal choice in terms of scoring rate is to respond with a ``Topspin'' stroke in the next play. Moving on to Figure~\ref{fig:case_2}(C), it becomes evident that following two strokes of the ``Serve'' and ``Short'' techniques, persisting with another ``Short'' stroke or responding with ``Others'' leads to a sudden drop in scoring rate to around $0.43$. This underscores that taking the initiative early in the game and launching an offensive increases our likelihood of winning. This observation is corroborated in other sequences with high scoring rate, where, in the majority of these sequences, the athlete who wins initiates offensive techniques earlier than his opponents. The discovered insights were confirmed by professional analysts collaborating with the Chinese national table tennis team.

Such analysis can be further applied to specific phases of a match and individual opponents, enabling the discovery of optimal tactical choices for each segment of the game against a particular player. This refined approach offers tangible benefits to both players and coaches in their tactical preparation, fostering a more nuanced understanding of the game's dynamics and enhancing competitive edge.

\section{Conclusion}
In this work, we propose a model, \system{}, to recognize and analyze stroke techniques in racket sports videos, facilitating automated tactical analysis. 
The fundamental insight lies in integrating sparse visual information with contextual domain knowledge to enhance high-level video understanding.
The efficacy of the proposed model is substantiated through a series of comparative experiments, ablation studies, and two case studies validated by analysts from the Chinese national table tennis team.

In the future, we envision extending this work further in two aspects. First, we plan to incorporate more nuanced context into the domain knowledge module, such as ball placement and player position. This enhancement aims to uncover optimal tactics tailored to specific contexts, adding another layer of sophistication to our analysis. Second, we intend to utilize the current technique-transition graph to unearth personalized tactical features, a step that promises to further enrich our discovery of insightful patterns.

\section{Acknowledgements}
The work was supported by NSF of China (U22A2032), Key ``Pioneer'' R\&D Projects of Zhejiang Province (2023C01120), and the Collaborative Innovation Center of Artificial Intelligence by MOE and Zhejiang Provincial Government (ZJU).

\bibliography{main}

\begin{thebibliography}{36}
\providecommand{\natexlab}[1]{#1}

\bibitem[{Behrmann et~al.(2022)Behrmann, Golestaneh, Kolter, Gall, and Noroozi}]{DBLP:conf/eccv/BehrmannGKGN22}
Behrmann, N.; Golestaneh, S.~A.; Kolter, Z.; Gall, J.; and Noroozi, M. 2022.
\newblock Unified Fully and Timestamp Supervised Temporal Action Segmentation via Sequence to Sequence Translation.
\newblock In Avidan, S.; Brostow, G.~J.; Ciss{\'{e}}, M.; Farinella, G.~M.; and Hassner, T., eds., \emph{Computer Vision - {ECCV} 2022 - 17th European Conference, Tel Aviv, Israel, October 23-27, 2022, Proceedings, Part {XXXV}}, volume 13695 of \emph{Lecture Notes in Computer Science}, 52--68. Springer.

\bibitem[{Bian et~al.(2022)Bian, Wang, Xiong, Huang, Liu, Li, Cheng, Zhao, Lu, and Dou}]{DBLP:journals/corr/abs-2207-12730}
Bian, J.; Wang, Q.; Xiong, H.; Huang, J.; Liu, C.; Li, X.; Cheng, J.; Zhao, J.; Lu, F.; and Dou, D. 2022.
\newblock P\({}^{\mbox{2}}\)A: {A} Dataset and Benchmark for Dense Action Detection from Table Tennis Match Broadcasting Videos.
\newblock \emph{CoRR}, abs/2207.12730.

\bibitem[{Carreira and Zisserman(2017)}]{DBLP:conf/cvpr/CarreiraZ17}
Carreira, J.; and Zisserman, A. 2017.
\newblock Quo Vadis, Action Recognition? {A} New Model and the Kinetics Dataset.
\newblock In \emph{2017 {IEEE} Conference on Computer Vision and Pattern Recognition, {CVPR} 2017, Honolulu, HI, USA, July 21-26, 2017}, 4724--4733. {IEEE} Computer Society.

\bibitem[{Chen et~al.(2020)Chen, Li, Bao, AlRegib, and Kira}]{DBLP:conf/cvpr/ChenLBAK20}
Chen, M.; Li, B.; Bao, Y.; AlRegib, G.; and Kira, Z. 2020.
\newblock Action Segmentation With Joint Self-Supervised Temporal Domain Adaptation.
\newblock In \emph{2020 {IEEE/CVF} Conference on Computer Vision and Pattern Recognition, {CVPR} 2020, Seattle, WA, USA, June 13-19, 2020}, 9451--9460. Computer Vision Foundation / {IEEE}.

\bibitem[{Chen et~al.(2022)Chen, Wei, Li, and Cai}]{DBLP:conf/cvpr/0001WL022}
Chen, M.; Wei, F.; Li, C.; and Cai, D. 2022.
\newblock Frame-wise Action Representations for Long Videos via Sequence Contrastive Learning.
\newblock In \emph{{IEEE/CVF} Conference on Computer Vision and Pattern Recognition, {CVPR} 2022, New Orleans, LA, USA, June 18-24, 2022}, 13791--13800. {IEEE}.

\bibitem[{Chu et~al.(2022)Chu, Xie, Ye, Lu, Xiao, Yuan, Chen, Zhang, and Wu}]{tivee2022chu}
Chu, X.; Xie, X.; Ye, S.; Lu, H.; Xiao, H.; Yuan, Z.; Chen, Z.; Zhang, H.; and Wu, Y. 2022.
\newblock TIVEE: Visual Exploration and Explanation of Badminton Tactics in Immersive Visualizations.
\newblock \emph{IEEE Transactions on Visualization and Computer Graphics}, 28(1): 118--128.

\bibitem[{Dai et~al.(2019)Dai, Xu, Yu, and Zhou}]{abd2019dai}
Dai, W.; Xu, Q.; Yu, Y.; and Zhou, Z. 2019.
\newblock Bridging Machine Learning and Logical Reasoning by Abductive Learning.
\newblock In Wallach, H.~M.; Larochelle, H.; Beygelzimer, A.; d'Alch{\'{e}}{-}Buc, F.; Fox, E.~B.; and Garnett, R., eds., \emph{Advances in Neural Information Processing Systems 32: Annual Conference on Neural Information Processing Systems 2019, NeurIPS 2019, December 8-14, 2019, Vancouver, BC, Canada}, 2811--2822.

\bibitem[{Deng et~al.(2021)Deng, Wu, Wang, Wu, Xie, Zhou, Zhang, Zhang, and Wu}]{DBLP:conf/chi/DengWWWXZZZW21}
Deng, D.; Wu, J.; Wang, J.; Wu, Y.; Xie, X.; Zhou, Z.; Zhang, H.; Zhang, X.~L.; and Wu, Y. 2021.
\newblock EventAnchor: Reducing Human Interactions in Event Annotation of Racket Sports Videos.
\newblock In Kitamura, Y.; Quigley, A.; Isbister, K.; Igarashi, T.; Bj{\o}rn, P.; and Drucker, S.~M., eds., \emph{{CHI} '21: {CHI} Conference on Human Factors in Computing Systems, Virtual Event / Yokohama, Japan, May 8-13, 2021}, 73:1--73:13. {ACM}.

\bibitem[{Farha and Gall(2019)}]{DBLP:conf/cvpr/FarhaG19}
Farha, Y.~A.; and Gall, J. 2019.
\newblock {MS-TCN:} Multi-Stage Temporal Convolutional Network for Action Segmentation.
\newblock In \emph{{IEEE} Conference on Computer Vision and Pattern Recognition, {CVPR} 2019, Long Beach, CA, USA, June 16-20, 2019}, 3575--3584. Computer Vision Foundation / {IEEE}.

\bibitem[{Huang et~al.(2019)Huang, Liao, Chen, Ik, and Peng}]{DBLP:conf/avss/HuangLCIP19}
Huang, Y.; Liao, I.; Chen, C.; Ik, T.; and Peng, W. 2019.
\newblock TrackNet: {A} Deep Learning Network for Tracking High-speed and Tiny Objects in Sports Applications.
\newblock In \emph{16th {IEEE} International Conference on Advanced Video and Signal Based Surveillance, {AVSS} 2019, Taipei, Taiwan, September 18-21, 2019}, 1--8. {IEEE}.

\bibitem[{Karpathy et~al.(2014)Karpathy, Toderici, Shetty, Leung, Sukthankar, and Fei{-}Fei}]{DBLP:conf/cvpr/KarpathyTSLSF14}
Karpathy, A.; Toderici, G.; Shetty, S.; Leung, T.; Sukthankar, R.; and Fei{-}Fei, L. 2014.
\newblock Large-Scale Video Classification with Convolutional Neural Networks.
\newblock In \emph{2014 {IEEE} Conference on Computer Vision and Pattern Recognition, {CVPR} 2014, Columbus, OH, USA, June 23-28, 2014}, 1725--1732. {IEEE} Computer Society.

\bibitem[{Kim, Kang, and Kim(2022)}]{DBLP:conf/eccv/KimKK22}
Kim, Y.~H.; Kang, H.; and Kim, S.~J. 2022.
\newblock A Sliding Window Scheme for Online Temporal Action Localization.
\newblock In Avidan, S.; Brostow, G.~J.; Ciss{\'{e}}, M.; Farinella, G.~M.; and Hassner, T., eds., \emph{Computer Vision - {ECCV} 2022 - 17th European Conference, Tel Aviv, Israel, October 23-27, 2022, Proceedings, Part {XXXIV}}, volume 13694 of \emph{Lecture Notes in Computer Science}, 653--669. Springer.

\bibitem[{Lea et~al.(2017)Lea, Flynn, Vidal, Reiter, and Hager}]{DBLP:conf/cvpr/LeaFVRH17}
Lea, C.; Flynn, M.~D.; Vidal, R.; Reiter, A.; and Hager, G.~D. 2017.
\newblock Temporal Convolutional Networks for Action Segmentation and Detection.
\newblock In \emph{2017 {IEEE} Conference on Computer Vision and Pattern Recognition, {CVPR} 2017, Honolulu, HI, USA, July 21-26, 2017}, 1003--1012. {IEEE} Computer Society.

\bibitem[{Lin et~al.(2020)Lin, Li, Wang, Tai, Luo, Cui, Wang, Li, Huang, and Ji}]{lin2020fast}
Lin, C.; Li, J.; Wang, Y.; Tai, Y.; Luo, D.; Cui, Z.; Wang, C.; Li, J.; Huang, F.; and Ji, R. 2020.
\newblock Fast learning of temporal action proposal via dense boundary generator.
\newblock In \emph{Proceedings of the AAAI conference on artificial intelligence}, 11499--11506.

\bibitem[{Lin et~al.(2023)Lin, Aouididi, Chen, Beyer, Pfister, and Wang}]{lin2023vird}
Lin, T.; Aouididi, A.; Chen, Z.; Beyer, J.; Pfister, H.; and Wang, J.-H. 2023.
\newblock VIRD: Immersive Match Video Analysis for High-Performance Badminton Coaching.
\newblock arXiv:2307.12539.

\bibitem[{Lin et~al.(2019)Lin, Liu, Li, Ding, and Wen}]{DBLP:conf/iccv/LinLLDW19}
Lin, T.; Liu, X.; Li, X.; Ding, E.; and Wen, S. 2019.
\newblock {BMN:} Boundary-Matching Network for Temporal Action Proposal Generation.
\newblock In \emph{2019 {IEEE/CVF} International Conference on Computer Vision, {ICCV} 2019, Seoul, Korea (South), October 27 - November 2, 2019}, 3888--3897. {IEEE}.

\bibitem[{Lin et~al.(2018)Lin, Zhao, Su, Wang, and Yang}]{DBLP:conf/eccv/LinZSWY18}
Lin, T.; Zhao, X.; Su, H.; Wang, C.; and Yang, M. 2018.
\newblock {BSN:} Boundary Sensitive Network for Temporal Action Proposal Generation.
\newblock In Ferrari, V.; Hebert, M.; Sminchisescu, C.; and Weiss, Y., eds., \emph{Computer Vision - {ECCV} 2018 - 15th European Conference, Munich, Germany, September 8-14, 2018, Proceedings, Part {IV}}, volume 11208 of \emph{Lecture Notes in Computer Science}, 3--21. Springer.

\bibitem[{Liu et~al.(2022)Liu, Wang, Wang, Ma, and Qiao}]{fineaction2022liu}
Liu, Y.; Wang, L.; Wang, Y.; Ma, X.; and Qiao, Y. 2022.
\newblock FineAction: A Fine-Grained Video Dataset for Temporal Action Localization.
\newblock \emph{IEEE Transactions on Image Processing}, 31: 6937--6950.

\bibitem[{Polk et~al.(2020)Polk, Jäckle, Häußler, and Yang}]{courttime2020tom}
Polk, T.; Jäckle, D.; Häußler, J.; and Yang, J. 2020.
\newblock CourtTime: Generating Actionable Insights into Tennis Matches Using Visual Analytics.
\newblock \emph{IEEE Transactions on Visualization and Computer Graphics}, 26(1): 397--406.

\bibitem[{Polk et~al.(2014)Polk, Yang, Hu, and Zhao}]{tennivis2014tom}
Polk, T.; Yang, J.; Hu, Y.; and Zhao, Y. 2014.
\newblock TenniVis: Visualization for Tennis Match Analysis.
\newblock \emph{IEEE Transactions on Visualization and Computer Graphics}, 20(12): 2339--2348.

\bibitem[{Shao et~al.(2020)Shao, Zhao, Dai, and Lin}]{DBLP:conf/cvpr/ShaoZDL20a}
Shao, D.; Zhao, Y.; Dai, B.; and Lin, D. 2020.
\newblock FineGym: {A} Hierarchical Video Dataset for Fine-Grained Action Understanding.
\newblock In \emph{2020 {IEEE/CVF} Conference on Computer Vision and Pattern Recognition, {CVPR} 2020, Seattle, WA, USA, June 13-19, 2020}, 2613--2622. Computer Vision Foundation / {IEEE}.

\bibitem[{Shou et~al.(2017)Shou, Chan, Zareian, Miyazawa, and Chang}]{DBLP:conf/cvpr/ShouCZMC17}
Shou, Z.; Chan, J.; Zareian, A.; Miyazawa, K.; and Chang, S. 2017.
\newblock {CDC:} Convolutional-De-Convolutional Networks for Precise Temporal Action Localization in Untrimmed Videos.
\newblock In \emph{2017 {IEEE} Conference on Computer Vision and Pattern Recognition, {CVPR} 2017, Honolulu, HI, USA, July 21-26, 2017}, 1417--1426. {IEEE} Computer Society.

\bibitem[{Singhania, Rahaman, and Yao(2021)}]{singhania2021coarse}
Singhania, D.; Rahaman, R.; and Yao, A. 2021.
\newblock Coarse to Fine Multi-Resolution Temporal Convolutional Network.
\newblock arXiv:2105.10859.

\bibitem[{Tong et~al.(2022)Tong, Song, Wang, and Wang}]{DBLP:journals/corr/abs-2203-12602}
Tong, Z.; Song, Y.; Wang, J.; and Wang, L. 2022.
\newblock VideoMAE: Masked Autoencoders are Data-Efficient Learners for Self-Supervised Video Pre-Training.
\newblock \emph{CoRR}, abs/2203.12602.

\bibitem[{Tran et~al.(2015)Tran, Bourdev, Fergus, Torresani, and Paluri}]{DBLP:conf/iccv/TranBFTP15}
Tran, D.; Bourdev, L.~D.; Fergus, R.; Torresani, L.; and Paluri, M. 2015.
\newblock Learning Spatiotemporal Features with 3D Convolutional Networks.
\newblock In \emph{2015 {IEEE} International Conference on Computer Vision, {ICCV} 2015, Santiago, Chile, December 7-13, 2015}, 4489--4497. {IEEE} Computer Society.

\bibitem[{Voeikov, Falaleev, and Baikulov(2020)}]{DBLP:conf/cvpr/VoeikovFB20}
Voeikov, R.; Falaleev, N.; and Baikulov, R. 2020.
\newblock TTNet: Real-time temporal and spatial video analysis of table tennis.
\newblock In \emph{2020 {IEEE/CVF} Conference on Computer Vision and Pattern Recognition, {CVPR} Workshops 2020, Seattle, WA, USA, June 14-19, 2020}, 3866--3874. Computer Vision Foundation / {IEEE}.

\bibitem[{Wang et~al.(2021{\natexlab{a}})Wang, Deng, Xie, Shu, Huang, Cai, Zhang, Zhang, Zhou, and Wu}]{DBLP:conf/kdd/WangDXSHCZZZW21}
Wang, J.; Deng, D.; Xie, X.; Shu, X.; Huang, Y.; Cai, L.; Zhang, H.; Zhang, M.; Zhou, Z.; and Wu, Y. 2021{\natexlab{a}}.
\newblock Tac-Valuer: Knowledge-based Stroke Evaluation in Table Tennis.
\newblock In Zhu, F.; Ooi, B.~C.; and Miao, C., eds., \emph{{KDD} '21: The 27th {ACM} {SIGKDD} Conference on Knowledge Discovery and Data Mining, Virtual Event, Singapore, August 14-18, 2021}, 3688--3696. {ACM}.

\bibitem[{Wang et~al.(2021{\natexlab{b}})Wang, Wu, Cao, Zhou, Zhang, and Wu}]{tacminer21wang}
Wang, J.; Wu, J.; Cao, A.; Zhou, Z.; Zhang, H.; and Wu, Y. 2021{\natexlab{b}}.
\newblock Tac-Miner: Visual Tactic Mining for Multiple Table Tennis Matches.
\newblock \emph{IEEE Transactions on Visualization and Computer Graphics}, 27(6): 2770--2782.

\bibitem[{Wang et~al.(2020)Wang, Zhao, Deng, Cao, Xie, Zhou, Zhang, and Wu}]{DBLP:journals/tvcg/WangZDCXZZW20}
Wang, J.; Zhao, K.; Deng, D.; Cao, A.; Xie, X.; Zhou, Z.; Zhang, H.; and Wu, Y. 2020.
\newblock Tac-Simur: Tactic-based Simulative Visual Analytics of Table Tennis.
\newblock \emph{{IEEE} Trans. Vis. Comput. Graph.}, 26(1): 407--417.

\bibitem[{Wang et~al.(2019)Wang, Xiong, Wang, Qiao, Lin, Tang, and Gool}]{DBLP:journals/pami/0002X00LTG19}
Wang, L.; Xiong, Y.; Wang, Z.; Qiao, Y.; Lin, D.; Tang, X.; and Gool, L.~V. 2019.
\newblock Temporal Segment Networks for Action Recognition in Videos.
\newblock \emph{{IEEE} Trans. Pattern Anal. Mach. Intell.}, 41(11): 2740--2755.

\bibitem[{Wu et~al.(2018)Wu, Lan, Shu, Ji, Zhao, Wang, and Zhang}]{DBLP:journals/tvcg/WuLSJZWZ18}
Wu, Y.; Lan, J.; Shu, X.; Ji, C.; Zhao, K.; Wang, J.; and Zhang, H. 2018.
\newblock iTTVis: Interactive Visualization of Table Tennis Data.
\newblock \emph{{IEEE} Trans. Vis. Comput. Graph.}, 24(1): 709--718.

\bibitem[{Xu et~al.(2020)Xu, Zhao, Rojas, Thabet, and Ghanem}]{xu2020g}
Xu, M.; Zhao, C.; Rojas, D.~S.; Thabet, A.; and Ghanem, B. 2020.
\newblock G-tad: Sub-graph localization for temporal action detection.
\newblock In \emph{Proceedings of the IEEE/CVF conference on computer vision and pattern recognition}, 10156--10165.

\bibitem[{Ye et~al.(2021)Ye, Chen, Chu, Wang, Fu, Shen, Zhou, and Wu}]{shuttle2021ye}
Ye, S.; Chen, Z.; Chu, X.; Wang, Y.; Fu, S.; Shen, L.; Zhou, K.; and Wu, Y. 2021.
\newblock ShuttleSpace: Exploring and Analyzing Movement Trajectory in Immersive Visualization.
\newblock \emph{IEEE Transactions on Visualization and Computer Graphics}, 27(2): 860--869.

\bibitem[{Yi, Wen, and Jiang(2021)}]{DBLP:conf/bmvc/YiWJ21}
Yi, F.; Wen, H.; and Jiang, T. 2021.
\newblock ASFormer: Transformer for Action Segmentation.
\newblock In \emph{32nd British Machine Vision Conference 2021, {BMVC} 2021, Online, November 22-25, 2021}, 236. {BMVA} Press.

\bibitem[{Zhang et~al.(2021)Zhang, Sciutto, Agrawala, and Fatahalian}]{zhang2021vid2player}
Zhang, H.; Sciutto, C.; Agrawala, M.; and Fatahalian, K. 2021.
\newblock Vid2player: Controllable video sprites that behave and appear like professional tennis players.
\newblock \emph{ACM Transactions on Graphics (TOG)}, 40(3): 1--16.

\bibitem[{Zhang et~al.(2023)Zhang, Yuan, Makoviychuk, Guo, Fidler, Peng, and Fatahalian}]{tennis2023zhang}
Zhang, H.; Yuan, Y.; Makoviychuk, V.; Guo, Y.; Fidler, S.; Peng, X.~B.; and Fatahalian, K. 2023.
\newblock Learning Physically Simulated Tennis Skills from Broadcast Videos.
\newblock \emph{ACM Trans. Graph.}, 42(4).

\end{thebibliography}

\end{document}